\documentclass[journal,twoside,hidelinks,a4paper]{IEEEtran}

\newcommand{\subparagraph}{}

\usepackage{emptypage,titling,titlesec,widetext,textcomp,relsize,parskip}
\usepackage[utf8]{inputenc}
\usepackage[margin=10pt,font=footnotesize,labelsep=period]{caption}
\usepackage{tabulary,longtable,makecell,multirow,color,colortbl}
\usepackage[colorlinks=false,linkcolor=black,urlcolor=gray,citecolor=black]{hyperref}
\usepackage[usenames,dvipsnames,svgnames,table]{xcolor}
\usepackage{amsmath,amssymb,amsfonts}
\usepackage{authblk,lipsum,txfonts,float,morefloats,gensymb,graphicx,setspace,fancyhdr,booktabs,array,marvosym,xhfill}
\usepackage[normalem]{ulem}
\usepackage[export]{adjustbox}
\usepackage[includeheadfoot,paperwidth=8.5in,paperheight=11in,top=1cm,bottom=2cm,left=2cm,right=2cm,footskip=.25in,headheight=40pt,headsep=12pt,heightrounded]{geometry}

\renewcommand{\figurename}{Figure}

\titlespacing\section{0pt}{8pt plus 4pt minus 2pt}{2pt plus 2pt minus 2pt}
\titlespacing\subsection{0pt}{8pt plus 4pt minus 2pt}{2pt plus 2pt minus 2pt}
\titlespacing\subsubsection{0pt}{8pt plus 4pt minus 2pt}{2pt plus 2pt minus 2pt}

\newcommand{\subtitle}[1]{\posttitle{\par\end{center}\begin{center}\bfseries\Large #1\end{center}\vskip1.2em}}
\newcommand\myabstract[2][.8]{\renewcommand\maketitlehookd{\mbox{}\medskip\par\centering\begin{minipage}{#1\textwidth}\small#2\end{minipage}}}

\definecolor{myred}{RGB}{189, 3, 34}
\date{}

\title{Information-Matched Auditing of Conversational Emotion-Shift Forecasting: A Leakage-Safe Evaluation Protocol}

\subtitle{}

\author[1]{\bfseries Bin Wen}
\author[2]{\bfseries Dai-Qiao Zhang}
\author[1]{\bfseries Tien-Ping Tan}

\affil[1]{School of Computer Sciences, Universiti Sains Malaysia, 11800 Gelugor, Penang, Malaysia}
\affil[2]{School of Mathematical Sciences, Universiti Sains Malaysia, 11800 Gelugor, Penang, Malaysia}
\affil[ ]{\vspace{0pt}}
\affil[ ]{Corresponding author: Tien-Ping Tan (e-mail: tienping@usm.my)}
\affil[ ]{\vspace{0pt}}

\begin{document}

\myabstract[.92]{\vspace{-1cm}\hrule\vspace{2mm}\textbf{Abstract:} Comparisons in conversational emotion-shift forecasting are easily distorted when a learned model receives rich features but its transition baseline is given a noisy or unavailable current-emotion label, when hyperparameter budgets differ, or when target-turn information enters the representation. We present an information-matched, leakage-safe auditing protocol that closes these gaps with an explicit commitment point, executable causality tests, equal four-configuration validation searches, and seed-dialogue-aware inference. Applying it to six causal model families on four text corpora and two feature-source sensitivity configurations shows that a deployable train-only current-emotion classifier feeding the transition baseline yields a different verdict than an oracle diagnostic where both sides receive gold current emotion. Under the deployable condition, learned models improve over the predicted-label baseline on all four primary corpora; DailyDialog gains are large and survive a 72-comparison Holm correction (best ROC-AUC gain = +0.167, adjusted $p=.0144$). Under matched gold information, the baseline remains competitive on IEMOCAP and stronger on MELD, while every model retains a large DailyDialog advantage. We additionally report precision-recall AUC, calibration, a strictly utterance-local feature control, same-person self-shift forecasting, and a future-access dose-response test. An eight-item leakage checklist accompanies the protocol. Code, raw seed predictions, and reporting artifacts are available from the corresponding author upon request.
\newline
\newline
\textbf{Keywords:} emotion forecasting, conversational emotion recognition, evaluation protocol, data leakage, benchmarking \vspace{1mm} \hrule}

\maketitle

\section{Introduction}
\label{sec:introduction}

Consider a call-center assistant that flags, one turn before it happens, that a customer is about to escalate from neutral to angry -- giving the agent seconds to change course. This is the promise of \emph{forecasting} emotion in conversation, as opposed to \emph{recognizing} it: the target state does not yet exist when the prediction is made. It is also precisely the setting in which a model can look right without being right. If any feature -- a pooled dialogue embedding, a bidirectional encoder, a context window one utterance too wide -- depends even slightly on the turn being forecast, the model is not anticipating the future; it is reading it off a crib sheet built into its own inputs. The failure is silent because nothing in a standard run signals it happened: loss curves look normal, F1 looks respectable, the leaderboard entry looks like progress.

Emotion recognition in conversation (ERC) has a large, mature literature (Section II). A smaller but growing thread reframes the task as forecasting -- predicting an upcoming emotional shift before the triggering utterance is seen, the setting relevant to anticipatory dialogue systems, de-escalation tools, and turn-taking agents. Forecasting is target-absent by construction: no feature may depend on utterance $n{+}1$. This constraint is easy to violate silently, e.g.\ via bidirectional encoders, dialogue-level pooling, or context windows not truncated at the decision point -- and the architectures ERC has spent a decade refining are, almost without exception, built to violate it, because recognition never had to enforce it in the first place.

This paper asks a measurement question: how do conclusions change when the future is actually hidden and the learned model and transition baseline receive comparable current-state information? The two requirements are distinct. A causal model without gold labels should be compared with a baseline driven by train-only predicted labels; a baseline supplied with gold $y_n$ should be compared with a model supplied with the same $y_n$. We contribute a measurement framework with six parts:

\begin{enumerate}
\item a formal commitment point and target-absent data construction for next-turn emotion shift;
\item an executable eight-item leakage audit plus future-perturbation causality tests;
\item train-only predicted-label and gold-label transition baselines that expose the information available in each comparison;
\item equal four-configuration validation searches and checkpoint selection for six causal model families, evaluated over three seeds;
\item seed--dialogue hierarchical intervals, paired dialogue-cluster permutation tests, and a declared familywise Holm correction; and
\item provenance, future-access, and speaker-target controls that test whether conclusions survive feature re-extraction, target redefinition, and controlled leakage.
\end{enumerate}

Applying this framework yields a picture that is conditional rather than uniformly negative: deployable models consistently improve on the predicted-label baseline, while an information-matched oracle baseline remains hard on several corpora. The final contribution matters because the two targets answer different questions. An interaction-level system may legitimately care whether the \emph{conversation's} next turn changes emotion regardless of who speaks. A coaching or wellbeing system may instead care whether the \emph{same person} changes state when they next speak. Reporting one as if it were the other can produce a semantically misleading claim even when the code is temporally leakage-free.

\section{Related Work}

\textbf{ERC backbones.} Our four ERC-derived models are representative of three context-modeling paradigms: recurrent interlocutor-state tracking \cite{majumder2019dialoguernn}, relational graph convolution \cite{ghosal2019dialoguegcn,ai2023dergcn}, and directed relational attention \cite{shen2021dagerc}. Knowledge-enriched encoders \cite{zhong2019ket,ghosal2020cosmic} and multimodal fusion architectures \cite{tsai2019mult,hu2021mmgcn,hazarika2020misa,li2023ga2mif,meng2024mglra} inform our feature choices (COSMIC RoBERTa; MM-DFN \cite{hu2021mmdfn}-style text+audio+visual). All are, in original form, bidirectional or graph-connected across the full dialogue -- appropriate for recognition, but a direct leakage source if repurposed for forecasting without a causal mask; interactive memory \cite{hazarika2018icon} and contextual-reasoning networks \cite{hu2021dialoguecrn} share this property.

\textbf{Emotion forecasting (EFC).} A parallel line explicitly targets an unobserved future utterance. Altarawneh et al.\ \cite{altarawneh2023pec} formalize PEC as predicting turn $n{+}1$ from context strictly up to $n$, decomposing the signal into sequence, self-dependency, and recency terms (the strategy we re-implement as PEC-style, Section V). Shi et al.\ \cite{shi2023epc,shi2024epc} model bilateral-speaker correlation and word-level cues in text+audio. Hi-EF \cite{wang2024hief} contributes a dedicated multimodal forecasting benchmark. MREPC \cite{ju2023mrepc} pre-trains contrastively over a short-term cross-speaker window. PUGCN \cite{xie2025pugcn} substitutes a pseudo-utterance-guided contrastive objective for the absent target (the strategy behind our Pseudo-future-style re-implementation). EDFC \cite{lu2023edfc} targets a next-utterance emotion distribution; ERFC \cite{debsharma2025erfc} shows feasibility in a deployed call-center setting. All are cross-speaker by design and, to our knowledge, none audits a leakage boundary as strict as ours.

\textbf{Emotion shift and flip: adjacent, not forecasting.} ESD-ERC \cite{gao2022esderc}, CFN-ESA \cite{li2024cfnesa}, and EmoShiftNet \cite{nirujan2025emoshiftnet} use shift labels to sharpen \emph{recognition} of an observed current utterance -- computed over turns the model can already see. Emotion-flip reasoning \cite{kumar2023efr,kumar2024ediref} runs backward: given a flip that has already occurred, it attributes the past trigger. Neither predicts an unobserved future shift under a leakage-safe protocol; we exclude both, noting our transition matrix is close in spirit to the inertia signal they are designed to overcome.

\textbf{Leakage and baseline validity.} Kapoor and Narayanan \cite{kapoor2023leakage} survey at least 294 papers across 17 fields and find leakage the dominant cause of irreproducible ML-based findings, proposing the taxonomy our 8-item audit is modeled after. Ferrari Dacrema et al.\ \cite{dacrema2019recsys} show most neural recommenders fail to beat tuned classical baselines once evaluation is corrected; Lin \cite{lin2019neuralhype} makes the same point for neural ranking. We view our transition-matrix result as the same phenomenon in emotion forecasting.

\section{Task Definition and Threat Model}
\label{sec:task}

Let a dialogue be $D=\{(x_t,y_t,s_t)\}_{t=1}^{T}$, where $x_t$ is an utterance-level feature vector, $y_t\in\{1,\ldots,K\}$ an emotion label, and $s_t$ a speaker identifier. At commitment point $n$, a legitimate forecaster may use only
\begin{equation}
\mathcal{I}_n=\{x_{1:n},s_{1:n}\}\quad\text{and, when declared, }y_n,
\label{eq:info}
\end{equation}
but no quantity computed from $x_{n+1:T}$ or $y_{n+1:T}$. This definition includes an important feature-provenance condition: $x_t$ must itself be utterance-local. Merely slicing a dialogue-level bidirectional representation at index $n$ does not make it causal.

\subsection{Two Forecasting Targets}

Our primary target is the \emph{interaction-level immediate shift}
\begin{equation}
z_n^{\mathrm{int}}=\mathbb{1}[y_{n+1}\neq y_n],\qquad n<T.
\label{eq:interaction}
\end{equation}
It is operationally appropriate when the question is ``will the emotion expressed in the next turn differ from the emotion expressed now?'' It is agnostic to speaker identity and therefore mixes two strata: $s_{n+1}=s_n$ and $s_{n+1}\neq s_n$.

For semantic clarity we additionally define the \emph{next-own-utterance self-shift}. Let $g(n)=\min\{m>n:s_m=s_n\}$ when such an utterance exists. Then
\begin{equation}
z_n^{\mathrm{self}}=\mathbb{1}[y_{g(n)}\neq y_n].
\label{eq:self}
\end{equation}
This target asks whether the current speaker changes emotion by the time they next speak, allowing intervening turns by others. It is not ``more correct'' in the abstract: it answers a different product and scientific question, has a variable forecast horizon, and excludes terminal appearances of each speaker. We report both rather than silently substituting one for the other.

\subsection{What Counts as Leakage?}

Table~\ref{tab:threats} distinguishes information leakage from other sources of optimistic or misleading evidence. The first three rows violate Eq.~\eqref{eq:info}; the remaining rows do not necessarily leak the future but can still make comparisons invalid.

\begin{table*}[t]
\centering
\footnotesize
\caption{Threat model for conversational emotion forecasting. The audit treats target-derived features and target-derived model states as zero-tolerance failures; the other threats require separate statistical or reporting controls.}
\label{tab:threats}
\begin{tabular}{p{0.17\textwidth}p{0.28\textwidth}p{0.24\textwidth}p{0.19\textwidth}}
\toprule
Threat & Failure mechanism & Detection & Required control \\
\midrule
Target-turn exposure & $x_{n+1}$ enters a window, pooled context, or auxiliary input & inspect window boundaries; sweep visible future turns & truncate at $n$ before any aggregation \\
Bidirectional state & output at $n$ depends on representations from $n+1{:}T$ & perturb future inputs and compare prefix logits & unidirectional recurrence or explicit causal mask \\
Feature provenance & nominally per-turn $x_n$ was extracted with dialogue-level future context & trace the feature extractor, not only the classifier & utterance-local extraction or causal re-extraction \\
Split contamination & duplicate dialogue text or the same global speaker occurs across folds & hash dialogue text; audit global identities & group-aware split and duplicate removal \\
Selection leakage & test fold determines threshold, epoch, or model family & inspect tuning log & validation-only selection with fixed test evaluation \\
Metric degeneracy & high base rate makes always-shift F1 look competitive & report prediction rate and balanced metrics & AUC, balanced accuracy, and constant baselines \\
Semantic target drift & cross-speaker next-turn change is described as same-person emotional change & stratify by $s_{n+1}=s_n$; evaluate Eq.~\eqref{eq:self} & name the target and deployment contract \\
\bottomrule
\end{tabular}
\end{table*}

\section{Protocol}

\textbf{Commitment point.} For decision point $n$, all input features (text/audio/visual) are restricted to utterances $\leq n$; the target label is $\mathbb{1}[y_{n+1} \neq y_n]$. All six learned models (Section V) are structurally causal, not causal-by-convention.

\textbf{Information-matched regimes.} In the \emph{deployable} regime, learned models receive utterance features and causal history but no gold emotion label. The transition baseline instead uses $\hat y_n$ from a linear current-emotion classifier fit exclusively on training utterances; its regularization is selected on the validation fold. In the \emph{oracle diagnostic}, the gold one-hot $y_n$ is appended to every learned model and the transition baseline conditions on the same gold $y_n$. This is not a deployment claim: it isolates the value of forecasting beyond a shared, perfect observation of current state.

\textbf{Causal construction and a zero-tolerance check.} DialogueRNN's global/party/emotion states update by chained \texttt{GRUCell} calls that never touch $x_{t+1:}$. DialogueGCN uses \emph{windowed} causal edges ($j{\to}i$ iff $j\leq i$ and $i-j\leq 6$); DAG-ERC uses \emph{unbounded} causal attention ($j\leq i$, any distance) -- two families differing sharply in receptive field yet still landing within the same $\lesssim$0.04 AUC band (Section V). We verified causality directly rather than trusting architecture alone: perturbing every input at $t\geq n$ to random noise leaves outputs at $t<n$ changed by exactly 0.0 (floating-point identical) for every model.

\textbf{Automated leakage audit.} Every experiment emits an 8-item checklist: last-utterance masking, model causality, per-utterance feature scope, speaker-split disjointness, val-only threshold tuning, dialogue-level bootstrap, train-fold-only baseline fitting, and baseline completeness.

\textbf{Strong baseline.} Beyond base-rate and no-change/inertia, we fit a speaker$\times$emotion transition matrix $P(y_{n+1} \mid y_n, \text{speaker})$ on the train fold only (Laplace $\alpha{=}1.0$; a speaker's row is used only with $\geq$20 training observations, else global backoff). This is the baseline every model must beat.

\textbf{Equal validation budget.} Each architecture receives the same four candidates spanning hidden size $\{64,128,256\}$, learning rate $\{3\!\times\!10^{-4},10^{-3}\}$, dropout $\{.1,.3\}$, and focal versus class-balanced cross-entropy loss. Seed 0 selects the configuration on validation ROC--AUC only; the selected configuration is then trained with seeds $\{0,1,2\}$ for at most 30 epochs, five-epoch patience after at least five epochs. The test fold is never evaluated during search or checkpoint selection.

\textbf{Metrics.} ROC--AUC is primary because it is threshold-free. We additionally report PR--AUC for the imbalanced shift class, Brier score and expected calibration error, shift-F1, precision, recall, and balanced accuracy. Predictors assigning more than 98\% of decisions to one class are automatically flagged.

\textbf{Significance.} Differences are estimated with a hierarchical bootstrap \cite{saravanan2020bootstrap} that first samples training seeds and then resamples whole test dialogues (1{,}999 replicates). Two-sided paired dialogue-cluster permutation tests use 4{,}999 sign flips and the plus-one correction, so a reported $p$-value is never zero. Holm--Bonferroni correction is applied jointly across the two information regimes, six models, and six dataset-configurations (72 declared comparisons).

\textbf{Reproducibility contract.} Random seeds are fixed at 0, 1, and 2. Each dataset is loaded and split before model construction; no fitted current-emotion classifier or cached prediction is shared across seeds. Dataset manifests record SHA-256 hashes, and independent tests verify that perturbing inputs at and after the forecast boundary cannot alter earlier logits. DailyDialog validation/test dialogues exactly duplicated in training are removed before any search.

\section{Experiments}
\label{sec:exp}

\textbf{Data.} The primary analysis uses four text corpora with released RoBERTa features (IEMOCAP \cite{busso2008iemocap}, MELD \cite{poria2019meld}, EmoryNLP \cite{zahiri2018emorynlp}, DailyDialog \cite{li2017dailydialog}; COSMIC pipeline \cite{ghosal2020cosmic}), plus two multimodal configurations (IEMOCAP, MELD with MM-DFN \cite{hu2021mmdfn} text+OpenSmile-audio+DenseNet-visual features) reported only as feature-source sensitivity analyses because the available release does not include the complete upstream extraction code. Table~\ref{tab:stats} gives size and shift-rate statistics (0.20--0.72) together with an F1-degeneracy analysis referenced below. The six configurations are not six independent corpora: IEMOCAP and MELD each appear once with COSMIC text features and once with MM-DFN multimodal features; inferential claims are made per configuration and are not treated as six independent replications. As an independent control that does not rely on released embeddings, we also fit TF--IDF vocabulary/IDF and truncated SVD on training utterances only and transform every utterance independently. DailyDialog contains 134 test and 101 validation dialogues exactly duplicated in training; all revised DailyDialog results remove those dialogues before configuration selection or evaluation.

\begin{table}[t]
\centering
\footnotesize
\caption{Dataset statistics and F1 degeneracy. Always-shift F1 $=2r/(1{+}r)$ for shift rate $r$ (analytic). Transition-BA $\approx$0.50 on MELD/EmoryNLP/IEMOCAP-mm shows standard F1 sits at the degenerate ceiling on exactly the corpora where shift is common; only IEMOCAP and DailyDialog clear it.}
\label{tab:stats}
\resizebox{\linewidth}{!}{%
\begin{tabular}{lrrrcc}
\toprule
Dataset & \#Dial & \#Utt & Shift $r$ & Always-shift F1 & Transition BA \\
\midrule
IEMOCAP     & 151    & 7{,}433   & 0.429 & 0.600 & 0.611 \\
MELD        & 1{,}432 & 13{,}708 & 0.581 & 0.735 & 0.500* \\
EmoryNLP    & 827    & 9{,}489   & 0.722 & 0.839 & 0.500* \\
DailyDialog & 13{,}118 & 102{,}979 & 0.198 & 0.331 & 0.681 \\
IEMOCAP-mm  & 151    & 7{,}433   & 0.429 & 0.600 & 0.500* \\
MELD-mm     & 1{,}432 & 13{,}708 & 0.581 & 0.735 & 0.500* \\
\bottomrule
\end{tabular}}
\end{table}

\subsection{Compared Models}

The benchmark spans recurrence, structured speaker state, graph propagation, directed attention, and two forecasting-specific strategies. Each model maps the same feature sequence to one logit per observed turn and uses the same loss and optimization budget. DialogueRNN maintains global and party states; causal DialogueGCN propagates messages only from earlier nodes in a six-turn window; DAG-ERC uses unbounded causal attention; PEC-style concatenates sequence, self-dependency, and exponentially weighted recency states; and Pseudo-future-style predicts an embedding for $x_{n+1}$ from the causal state before classifying shift, where the actual $x_{n+1}$ is only an auxiliary regression target during training and never a forward-path input.

\textbf{State-factorized control.} Motivated by work that predicts the next-utterance emotion distribution directly \cite{lu2023edfc}, we add a causal multi-task control with three heads: current emotion, future emotion, and direct shift. It reaches .628/.628/.546/.747 ROC--AUC on IEMOCAP/MELD/EmoryNLP/DailyDialog and improves over the deployable transition baseline by +.052/+.041/+.043/+.166; MELD and DailyDialog remain significant after four-test Holm correction ($p=.0036$ and $.0008$). We also retain a bidirectional GRU solely as a positive leakage control (not a candidate forecaster): if the leaky model improves sharply while the safe model does not, this argues against a failure of the benchmark implementation itself.

\subsection{Speaker-Relation Audit}
\label{sec:speaker}

\begin{figure*}[t]
\centering
\includegraphics[width=0.85\textwidth]{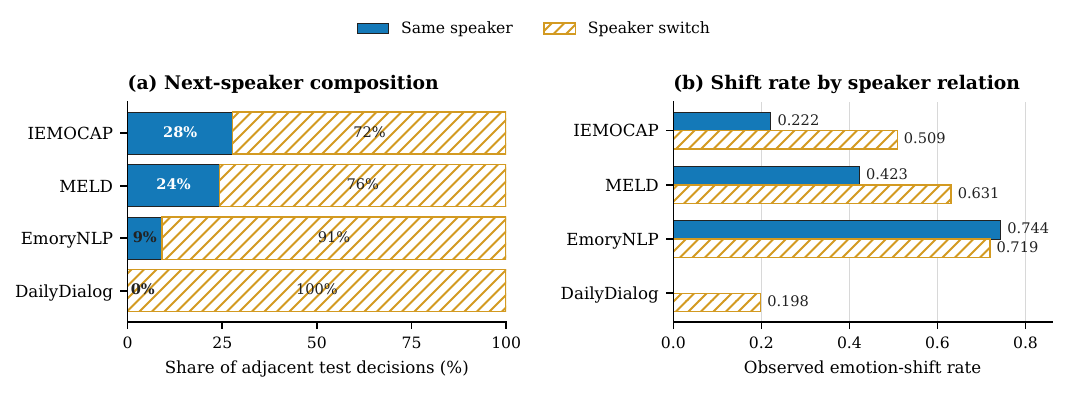}
\caption{Speaker composition and emotion-shift rate at adjacent test decision points. Filled bars denote same-speaker transitions; hatched bars denote speaker switches. DailyDialog has no adjacent same-speaker decisions. Exact denominators are given in Table~\ref{tab:speaker}.}
\label{fig:speaker}
\end{figure*}

Table~\ref{tab:speaker} exposes a compositional property hidden by aggregate shift rates. In the test folds, only 27.8\% of IEMOCAP and 24.2\% of MELD adjacent decisions retain the same speaker; EmoryNLP retains the speaker in 9.1\%, and DailyDialog alternates speakers by construction. Most labels in Eq.~\eqref{eq:interaction} therefore compare the emotions of two different people. IEMOCAP's cross-speaker shift rate is 0.509 versus 0.222 for same-speaker adjacent turns; MELD's is 0.631 versus 0.423 (Figure~\ref{fig:speaker}). A 5{,}000-sample dialogue-cluster bootstrap confirms the switch-minus-same difference is not an utterance-count artifact on IEMOCAP ($+0.287$, 95\% CI $[+0.200,+0.373]$) or MELD ($+0.208$, $[+0.151,+0.265]$); both two-sided $p<.001$. EmoryNLP shows no supported difference ($-0.025$, $p=.614$).

\begin{table*}[t]
\centering
\footnotesize
\caption{Speaker relation and target composition on the four unique test corpora. Same-next is $P(s_{n+1}=s_n)$. Self-shift uses the next-own-utterance target in Eq.~\eqref{eq:self}; gap is the mean number of turns from $n$ to $g(n)$.}
\label{tab:speaker}
\begin{tabular}{lrrrrrrr}
\toprule
Corpus & Immediate $N$ & Same-next & Shift$\mid$same & Shift$\mid$switch & Self $N$ & Self shift & Mean gap \\
\midrule
IEMOCAP     & 1{,}592 & 0.278 & 0.222 & 0.509 & 1{,}561 & 0.263 & 1.98 \\
MELD        & 2{,}330 & 0.242 & 0.423 & 0.631 & 1{,}864 & 0.538 & 2.21 \\
EmoryNLP    & 905     & 0.091 & 0.744 & 0.719 & 739     & 0.673 & 2.51 \\
DailyDialog & 6{,}740 & 0.000 & --    & 0.198 & 5{,}740 & 0.180 & 2.00 \\
\bottomrule
\end{tabular}
\end{table*}

\begin{table}[t]
\centering
\footnotesize
\caption{Train-fold transition baselines under the two targets (test AUC). These are label-only diagnostics, not learned-model results.}
\label{tab:selfbase}
\begin{tabular}{lcccc}
\toprule
& \multicolumn{2}{c}{Immediate} & \multicolumn{2}{c}{Self-shift} \\
Corpus & Global & Speaker & Global & Speaker \\
\midrule
IEMOCAP     & 0.642 & 0.642 & 0.586 & 0.586 \\
MELD        & 0.676 & 0.676 & 0.663 & 0.663 \\
EmoryNLP    & 0.608 & 0.608 & 0.592 & 0.592 \\
DailyDialog & 0.691 & 0.691 & 0.657 & 0.657 \\
\bottomrule
\end{tabular}
\end{table}

Speaker conditioning contributes exactly zero AUC on every corpus: test-time speaker keys are, by construction, absent from the train-fold speaker table, so the estimator always backs off to the global row. This evaluation provides no evidence that speaker-identity conditioning helps, only that dialogue-local role identities cannot be tested for a within-person effect without a genuinely cross-dialogue speaker identifier. Forecast horizon is another hidden source of heterogeneity (Table~\ref{tab:horizon}): IEMOCAP and MELD show markedly higher shift rates after gaps of at least three turns than at gap one.

\begin{table}[t]
\centering
\footnotesize
\caption{Self-shift composition by forecast gap. Cells show test $N$ / shift rate; -- indicates the released test split contains no decision at that gap.}
\label{tab:horizon}
\begin{tabular}{lccc}
\toprule
Corpus & Gap 1 & Gap 2 & Gap $\geq3$ \\
\midrule
IEMOCAP     & 442 / .222 & 864 / .234 & 255 / .431 \\
MELD        & 565 / .423 & 961 / .580 & 338 / .612 \\
EmoryNLP    & 82 / .744  & 464 / .619 & 193 / .772 \\
DailyDialog & --         & 5{,}740 / .180 & -- \\
\bottomrule
\end{tabular}
\end{table}

\begin{table*}[t]
\centering
\footnotesize
\caption{Complete next-own-utterance self-shift ROC--AUC (three-seed mean). Trans-Pred is information-matched. DailyDialog gains all survive the separate 24-comparison Holm correction ($p_{\rm adj}=.0048$).}
\label{tab:selfmodels}
\begin{tabular}{lccccccc}\toprule
Corpus & Trans-Pred & GRU & DRNN & DGCN & DAG-ERC & PEC & PseudoFut \\
\midrule
IEMOCAP     & .541 & .609 & .602 & .597 & .598 & .593 & .605 \\
MELD        & .584 & .593 & .600 & .608 & .610 & .593 & .593 \\
EmoryNLP    & .518 & .509 & .503 & .514 & .515 & .501 & .523 \\
DailyDialog & .577 & \textbf{.699} & \textbf{.706} & \textbf{.698} & \textbf{.691} & \textbf{.695} & \textbf{.693} \\
\bottomrule
\end{tabular}
\end{table*}

\textbf{Main result (Table~\ref{tab:main}).} Information matching changes the conclusion. In the deployable regime, all six models exceed the train-only predicted-label transition baseline on every primary text corpus in point estimate. The gain is decisive on duplicate-free DailyDialog ($+0.160$ to $+0.167$ ROC--AUC; all adjusted $p=.0144$) and smaller elsewhere. Among the primary text corpora, DialogueRNN on MELD is the only other comparison that survives the 72-test familywise correction ($\Delta=+0.041$, adjusted $p=.0352$); all six MELD-mm gains also survive but sit in a feature-source sensitivity configuration rather than a primary corpus.

The oracle diagnostic (Table~\ref{tab:oracle}) asks a different question. When both sides receive gold $y_n$, all six models remain clearly better on DailyDialog ($+0.106$ to $+0.132$; adjusted $p=.0144$), whereas the transition baseline is stronger on MELD ($-0.032$ to $-0.064$), with four model differences surviving correction. IEMOCAP is effectively tied and EmoryNLP is directionally favorable to the baseline. Thus, the transition matrix is neither a universal lower bound nor a straw baseline: its strength depends on whether current emotion is known and on corpus-specific transition structure.

\begin{table*}[t]
\centering
\footnotesize
\caption{Deployable shift ROC--AUC (three-seed mean). Trans-Pred uses current-emotion predictions from a train-only linear classifier; learned models receive no gold emotion. Bold values exceed the matched baseline. DailyDialog is duplicate-free; MM rows are feature-source sensitivity analyses.}
\label{tab:main}
\begin{tabular}{lccccccc}\toprule
Dataset & Trans-Pred & GRU & DRNN & DGCN & DAG-ERC & PEC & PseudoFut \\
\midrule
IEMOCAP      & .576 & \textbf{.647} & \textbf{.627} & \textbf{.609} & \textbf{.607} & \textbf{.632} & \textbf{.631} \\
MELD         & .588 & \textbf{.621} & \textbf{.629} & \textbf{.623} & \textbf{.608} & \textbf{.620} & \textbf{.600} \\
EmoryNLP     & .501 & \textbf{.521} & \textbf{.533} & \textbf{.536} & \textbf{.554} & \textbf{.539} & \textbf{.542} \\
DailyDialog  & .580 & \textbf{.746} & \textbf{.744} & \textbf{.748} & \textbf{.744} & \textbf{.743} & \textbf{.741} \\
IEMOCAP-mm   & .566 & .560 & \textbf{.598} & \textbf{.593} & .559 & \textbf{.581} & \textbf{.578} \\
MELD-mm      & .584 & \textbf{.723} & \textbf{.716} & \textbf{.724} & \textbf{.720} & \textbf{.720} & \textbf{.723} \\
\bottomrule
\end{tabular}
\end{table*}

\begin{table*}[t]
\centering
\footnotesize
\caption{Oracle information-matched shift ROC--AUC (three-seed mean). Both the transition baseline and every learned model receive gold current emotion. Bold values exceed Trans-Gold.}
\label{tab:oracle}
\begin{tabular}{lccccccc}\toprule
Dataset & Trans-Gold & GRU & DRNN & DGCN & DAG-ERC & PEC & PseudoFut \\
\midrule
IEMOCAP      & .642 & \textbf{.649} & .615 & .617 & .616 & .631 & .610 \\
MELD         & .676 & .630 & .629 & .628 & .644 & .632 & .612 \\
EmoryNLP     & .608 & .553 & .561 & .561 & .563 & .591 & .525 \\
DailyDialog  & .694 & \textbf{.808} & \textbf{.820} & \textbf{.826} & \textbf{.822} & \textbf{.804} & \textbf{.800} \\
IEMOCAP-mm   & .642 & .566 & .590 & .598 & .577 & .594 & .569 \\
MELD-mm      & .676 & \textbf{.802} & \textbf{.791} & \textbf{.805} & \textbf{.807} & \textbf{.798} & \textbf{.798} \\
\bottomrule
\end{tabular}
\end{table*}

Tables~\ref{tab:main} and~\ref{tab:oracle} average three independently trained seeds; inference uses all three raw score vectors rather than a single representative run. On DailyDialog every model has a positive delta in every seed, the 95\% hierarchical intervals exclude zero, and all paired permutation tests remain significant after the declared joint correction.

\textbf{Independent utterance-local features.} Table~\ref{tab:feat} removes dependence on released deep embeddings. TF--IDF vocabulary and IDF, followed by rank-128 truncated SVD, are fitted only on training utterances. The local-feature GRU remains substantially above its matched predicted-label transition baseline on IEMOCAP and DailyDialog and directionally above it on EmoryNLP; MELD is tied. Crucially, the DailyDialog advantage is unchanged in magnitude ($+.164$ versus $+.166$ with RoBERTa), so its headline improvement cannot be attributed solely to unobserved context in an upstream encoder.

\begin{table}[t]
\centering
\footnotesize
\caption{Strict utterance-local feature control (three-seed ROC--AUC). Both the GRU features and the baseline's current-emotion classifier are refit from the same train-only TF--IDF/SVD rows.}
\label{tab:feat}
\begin{tabular}{lccc}
\toprule
Corpus & GRU & Trans-Pred & $\Delta$ [95\% CI] \\
\midrule
IEMOCAP     & .642 & .552 & +.089 [.010,.154] \\
MELD        & .539 & .538 & +.001 [-.027,.028] \\
EmoryNLP    & .555 & .494 & +.061 [.014,.112] \\
DailyDialog & .673 & .509 & +.164 [.141,.185] \\
\bottomrule
\end{tabular}
\end{table}

\textbf{Stronger predicted-label baseline.} A posterior-marginalized control computes $P(\mathrm{shift}\mid x_n)=\sum_k P(y_n{=}k\mid x_n)\,P(y_{n+1}{\neq}k\mid y_n{=}k,\text{speaker})$ instead of committing to the classifier's single most likely label. Marginalization is generally, though not uniformly, stronger: duplicate-free DailyDialog rises from .580 to .662 and MELD-mm from .584 to .605, while IEMOCAP dips slightly (.576 to .569). Every DailyDialog and MELD-mm model gain nevertheless remains corrected-significant in a separately declared 36-comparison Holm family ($+.078$ to $+.085$ and $+.111$ to $+.118$, respectively); MELD's smaller gains and EmoryNLP's vanish. The control sharpens rather than overturns the conditional conclusion.

\begin{figure*}[t]
\centering
\includegraphics[width=0.85\textwidth]{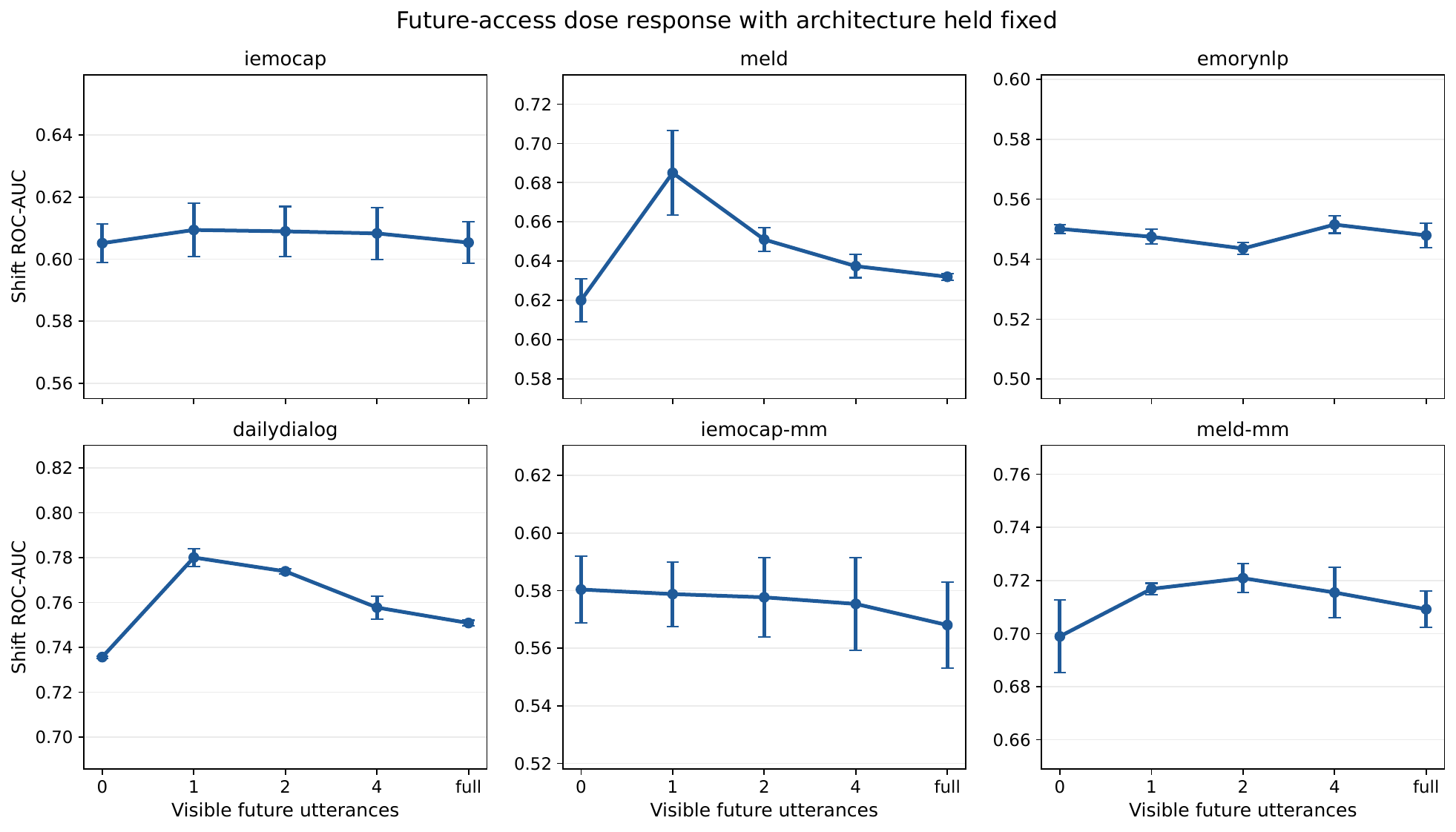}
\caption{Future-access dose response with one fixed Transformer architecture. Error bars show three-seed standard deviation. MELD and DailyDialog peak when the target utterance alone becomes visible; none of the six configurations supports a general monotonic-more-future claim.}
\label{fig:dose}
\end{figure*}

\textbf{Same-architecture dose response (Figure~\ref{fig:dose}).} A single-layer Transformer holds parameters, positional encoding, pooling, optimizer, and checkpoint rule fixed; only its attention mask changes so that position $n$ sees $k\in\{0,1,2,4,\mathrm{all}\}$ future utterances. A single layer is required for $k$ to mean what the sweep claims, since a stacked multi-layer encoder with the same window mask gives an effective receptive field of ${\sim}Lk$ for $L$ layers rather than $k$. The response is not monotone: MELD rises from .620 at $k=0$ to .685 at $k=1$, then falls to .651/.637/.632; duplicate-free DailyDialog similarly rises from .736 to .780 and falls to .774/.758/.751. The other four configurations vary by at most .022 without a consistent direction. Exposure to the target utterance can materially inflate performance, but additional future context need not add signal and may dilute optimization.

\subsection{Split and Leakage Robustness}
\label{sec:robust}

We hashed each dialogue by its complete ordered utterance text and compared validation/test hashes against the training fold (Table~\ref{tab:duplicates}). IEMOCAP and EmoryNLP contain no exact cross-fold matches; MELD contains one duplicate test dialogue; DailyDialog contains 134/1{,}000 duplicate test dialogues and 101/1{,}000 duplicate validation dialogues. We remove every matching DailyDialog validation/test dialogue before any configuration search, checkpoint selection, baseline fitting, or reported evaluation; test dialogues fall from 1,000 to 866 and validation dialogues from 1,000 to 899. All DailyDialog results in this paper use this same clean split.

\begin{table}[t]
\centering
\footnotesize
\caption{Exact train-fold dialogue duplicates. A duplicate has an identical full ordered text sequence, not merely an overlapping utterance.}
\label{tab:duplicates}
\begin{tabular}{lrrrr}
\toprule
& \multicolumn{2}{c}{Test} & \multicolumn{2}{c}{Validation} \\
Corpus & $N$ & Dup. & $N$ & Dup. \\
\midrule
IEMOCAP     & 31    & 0   & 12    & 0 \\
MELD        & 280   & 1   & 114   & 0 \\
EmoryNLP    & 79    & 0   & 89    & 0 \\
DailyDialog & 1{,}000 & 134 & 1{,}000 & 101 \\
\bottomrule
\end{tabular}
\end{table}

\section{Findings and Discussion}

The revised evidence supports six bounded findings: (1) information matching is consequential -- no-label models consistently exceed a baseline fed noisy train-only label predictions in point estimate, whereas a gold-label transition matrix is competitive or stronger on several corpora; (2) DailyDialog is the most stable positive case, with every tested model beating the hard predicted-label, posterior-marginalized, and gold-label baselines after duplicate removal, across every seed, with corrected-significant hierarchical inference; (3) MELD is conditional -- learned models improve on the deployable baseline but most lose to the oracle baseline, so the current-state observation, not only architecture, changes the answer; (4) ranking and probability quality diverge, since PR--AUC corroborates DailyDialog discrimination while Brier/ECE show that learned scores require calibration before risk interpretation; (5) feature provenance remains part of the estimand, as the independently fitted local-feature control preserves large DailyDialog and IEMOCAP gains but not the MELD gain; and (6) adjacent-turn shift is predominantly cross-speaker in all four corpora, so interaction-level next-turn shift and a person's next-own-utterance self-shift are distinct targets evaluated separately here.

These findings do not establish that one architecture class is generally superior or irrelevant. They establish a narrower measurement result: the sign and magnitude of model-versus-transition differences depend on the information supplied to both sides, and conclusions that omit this factor can reverse. The multimodal rows reinforce the need for restraint -- their point estimates can be large, but the complete upstream producers are unavailable, so they are useful sensitivity evidence rather than a basis for claiming a modality is intrinsically better.

\subsection{Operational Implications}

Temporal validity is necessary but not sufficient for deployment. A high-AUC alert can still be poorly calibrated, too late for intervention, or unfairly concentrated on particular speakers. The commitment point must correspond to a real action window: after the current turn is available but before any acoustic, lexical, or metadata trace of the next turn enters the system. Thresholds should follow intervention costs rather than F1 alone: in de-escalation, a false negative may miss harm while repeated false positives create alert fatigue. This distinction is visible in the strongest ranking result: duplicate-free DailyDialog DialogueGCN reaches ROC--AUC .748 and PR--AUC .441, but its uncalibrated Brier score/ECE are .188/.224, versus .153/.052 for the weaker predicted-label transition baseline. Post-hoc calibration must therefore be fitted on held-out deployment-like data before probabilities are interpreted as risks. This work audits measurement; it does not establish that an automated emotional intervention is safe or appropriate.

\subsection{Threats to Validity and Limitations}

\textbf{Construct and internal validity.} Emotion annotations are discrete expressions, not direct measures of internal affect. Causality tests cover model forward paths, masks, and provided feature arrays, but cannot retroactively certify every upstream feature-extraction step; the train-only TF--IDF/SVD control reduces, but cannot remove, this limitation.

\textbf{Split validity.} We remove exact train--test dialogue duplication from every DailyDialog run, but not subtler writing-style leakage. Only IEMOCAP's session split provides an unseen-actor test; speaker conditioning elsewhere means local role conditioning, not person-level generalization.

\textbf{Statistical validity.} Three seeds characterize optimization variability but do not make the six configurations independent, since two corpora recur under different feature sources; the joint Holm correction is intentionally conservative.

\textbf{Model scope.} Forecasting-specific systems are strategy-level re-implementations because original code was unavailable; the common four-configuration search favors comparability over per-model best-case tuning.

\textbf{Information matching is imperfect in the oracle direction.} In the oracle diagnostic, the gold one-hot label is concatenated to the per-turn feature vector at every position $t \le n$, so a sequential model's representation at $n$ is conditioned on the entire observed label history, not on $y_n$ alone; the transition baseline conditions only on the single most recent label. ``Information-matched'' therefore means matched at current-emotion availability, not in the total label information reaching each side, and the oracle gap likely understates the transition baseline's competitiveness under a strict single-label match. The symmetric deployable-side concern -- that a hard current-label prediction discards classifier uncertainty a stronger baseline could exploit -- is addressed by the posterior-marginalized control above.

\section{Minimal Reporting Checklist}

We distill a standalone checklist for future EFC-forecasting papers: (1) state the commitment point and verify no feature crosses it; (2) name the semantic target -- interaction-level next turn or same-person next own turn -- and report speaker-relation composition; (3) verify causality directly (e.g.\ a future-perturbation test), not by architecture claim alone; (4) trace feature provenance upstream of the classifier; (5) match current-label information between learned models and train-only transition baselines; (6) report ROC--AUC, PR--AUC, calibration, balanced accuracy, and prediction rate rather than F1 alone; (7) use seed--dialogue-aware paired inference corrected jointly across the declared comparison family; (8) label leaky ablations and exclude them from headline comparisons; (9) check duplicate dialogues and global speaker overlap; and (10) release masks, split identifiers, raw seed-level predictions, and audit logs.

\section{Conclusion}

Information-matched evaluation changes the interpretation of conversational emotion-shift forecasting. Models that receive no gold current label can meaningfully outperform a transition baseline driven by train-only predicted labels; on duplicate-free DailyDialog this conclusion is large, seed-consistent, and familywise significant. A transition matrix supplied with gold current emotion is nevertheless competitive on IEMOCAP and stronger on MELD, showing why privileged-label diagnostics must not be presented as deployable comparisons. Independent local features preserve the DailyDialog result, while calibration metrics warn that strong ranking is not yet reliable risk estimation. Finally, adjacent interaction shift and same-person self-shift are both legitimate but different targets requiring explicit horizons and matched baselines. The general lesson is methodological rather than architectural: state the temporal, information, provenance, and semantic contracts before interpreting a model comparison \cite{kapoor2023leakage,dacrema2019recsys,lin2019neuralhype}. Code, raw seed predictions, tests, and reporting artifacts supporting reproducible auditing are available from the corresponding author upon request.

\section*{Acknowledgment}

The authors used a large language model assistant to help with English-language editing, LaTeX formatting, code review, and the documentation and presentation of supplementary analyses. All experimental designs, executions, interpretations, and conclusions were independently reviewed and verified by the authors, who assume full responsibility for the submitted work.

\end{document}